%% file: output.tex
\documentclass[sigconf]{acmart}
\usepackage{algorithm}
\usepackage{algpseudocode}
\usepackage{amsmath}
\usepackage{amsmath}
\usepackage{amssymb}
\usepackage{booktabs}
\usepackage{color}
\usepackage{comment}
\usepackage{cuted}
\usepackage[inline]{enumitem}
\usepackage{epsfig}
\usepackage{etoolbox}
\usepackage{fancybox}
\usepackage{float}
\usepackage{graphicx}
\usepackage{hyperref}
\usepackage{listings}
\usepackage{lipsum}
\usepackage{multirow}
\usepackage{multicol}
\usepackage{makecell}
\usepackage{placeins}
\usepackage{rotating}
\usepackage[dvipsnames]{xcolor}
\usepackage{setspace}
\usepackage{url}
\usepackage{algorithm}
\usepackage{algpseudocode}
\usepackage{colortbl} 
\usepackage{ulem} 
\usepackage{numprint} 
\usepackage{fancyhdr} 
\usepackage{pifont} 
\usepackage[inline]{enumitem}
\usepackage{verbatim}
\usepackage{wrapfig}
\usepackage{bbding}

\begin{document}

\title{\textit{KERV}: Kinematic-Rectified Speculative Decoding for Embodied VLA Models}

\author{Zihao Zheng$^{1}$, Zhihao Mao$^{2}$, Maoliang Li$^{1}$, Jiayu Chen$^{1}$, Xinhao Sun$^{1}$, Zhaobo Zhang$^{1}$, Donggang Cao$^{1}$, Hong Mei$^{1}$, $^{\dagger}$Xiang Chen$^{1}$}
\affiliation{%
  \institution{$^{1}$ School of Computer Science, Peking University, Beijing, China \\
  $^{2}$ School of Computer Science, China University of Geosciences (Wuhan), Wuhan, China}
  \country{}}

\copyrightyear{2026}
\acmYear{2026}
\setcopyright{cc}
\setcctype{by-nc-nd}
\acmConference[DAC '26]{63rd ACM/IEEE Design Automation Conference}{July 26--29, 2026}{Long Beach, CA, USA}
\acmBooktitle{63rd ACM/IEEE Design Automation Conference (DAC '26), July 26--29, 2026, Long Beach, CA, USA}
\acmDOI{10.1145/3770743.3804238}
\acmISBN{979-8-4007-2254-7/2026/07}

\input{_tex/0_abstract.tex}

\maketitle

\begingroup
\renewcommand\thefootnote{}
\footnotetext{$^{\dagger}$ Corresponding author: Xiang Chen, $<$xiang.chen@pku.edu.cn$>$}
\endgroup

\input{_tex/1_introduction.tex}
\input{_tex/2_background.tex}
\input{_tex/3_motivation.tex}
\input{_tex/4_method.tex}
\input{_tex/5_implementation.tex}
\input{_tex/6_experiment.tex}
\input{_tex/7_conclusion.tex}
\input{_tex/8_ack.tex}

\clearpage
\balance
\bibliographystyle{ACM-Reference-Format.bst}
\bibliography{reference/ref.bib}

\end{document}

%% file: _tex/0_abstract.tex
\begin{abstract}

Vision-Language-Action (VLA) models build a token-domain robot control paradigm, yet suffer from low speed.
Speculative Decoding (SD) is an optimization strategy that can boost inference speed.
Two key issues emerge when integrating VLA and SD: first, SD relies on re-inference to address token errors, which is computationally expensive; second, to mitigate token errors, the acceptance threshold in SD requires careful adjustment.
Existing works fail to address the above two issues effectively.
Meanwhile, as the bridge between AI and the physical world, existing embodied intelligence has overlooked the application of robotic kinematics.
To address these issues, we innovatively combine token-domain VLA models with kinematic-domain prediction for SD, proposing a kinematic-rectified SD framework named \textit{KERV}.
We employ a kinematics-based Kalman Filter to predict actions and compensate for SD errors, avoiding costly re-inference.
Moreover, we design a kinematics-based adjustment strategy to dynamically rectify the acceptance threshold, addressing the difficulty of threshold determination.
Experimental results across diverse tasks and environments demonstrate that \textit{KERV} achieves 27\%$\sim$37\% acceleration with nearly no Success Rate loss.
\end{abstract}

%% file: _tex/1_introduction.tex
\section{Introduction}
\label{tex_introduction}

Vision-Language-Action (VLA) models have emerged as the mainstream for embodied intelligence~\cite{openvla, rt2}.
They propose a \textbf{token-domain paradigm} for robot control that unifies visual, textual, and action information as tokens, yet suffers from low inference speed~\cite{vla-survey, vla-survey-2, openhelix}.
Existing work accelerates VLA model inference via three key avenues: model architecture innovation~\cite{robomamba, tinyvla, edgevla, fastv}; model compression (pruning~\cite{efficient-vla, specprune-vla}, caching~\cite{vla-cache, think2act1}, quantization~\cite{QAIL, mbq}); and runtime optimizations (layer-skipping~\cite{mole-vla}, early-exit~\cite{deer-vla, ceed-vla} strategies).

Speculative Decoding (SD) is a promising decoding optimization technique that can effectively accelerate Large Language Model (LLM) inference speed.
Specifically, it uses a draft model to rapidly generate tokens and then efficiently verifies in parallel by LLMs~\cite{first-sd, draft-and-verify}. 
However, adapting SD to VLA models is non-trivial.
First, SD relies on re-inference to address the token errors during VLA's action generation, which leads to a high computational cost and limits the speed.
Second, to mitigate token errors, the acceptance threshold in SD requires careful adjustment to achieve an optimal trade-off between inference speed and task accuracy.
Existing works such as Spec-VLA~\cite{spec-vla} adopt relaxed acceptance to mitigate token errors, yet they set a static acceptance threshold, ignoring the complex variations in tasks and environments. 
Moreover, Spec-VLA also needs re-inference when facing token errors, which limits its performance.

Meanwhile, although embodied intelligence is widely recognized as the bridge linking AI to the physical world, most existing solutions fail to acknowledge the fundamental importance of kinematics.
Traditional robotic control~\cite{kinematics-1, kinematics-2} adopts a \textbf{kinematic-domain paradigm}, utilizing kinematic information (e.g., joint angles, link displacements, motion velocities) to directly regulate robot movements via pre-defined mathematical models and control laws. 
It relies on accurate robot dynamic modeling and real-time sensor feedback to achieve precise trajectory tracking and motion execution.
Unlike VLA models, it cannot conduct long-term task planning or intelligent environmental perception, but enables accurate motion control in shorter action contexts with lower computational complexity and high efficiency.
\textbf{This inspires us to combine the token-domain VLA paradigm with the kinematic-domain traditional control paradigm, leveraging the strengths of both.}

In this paper, we take SD optimization as the starting point, innovatively combine the VLA model with a kinematic-based method, and propose a kinematic-rectified SD framework termed \textit{KERV}.
(1) First, we analyze VLA model generation errors and acceptance thresholds from both token and kinematic perspectives, finding kinematic-based methods suitable as an adjustment and compensation strategy during VLA generation.
(2) Second, we propose a kinematic-based compensation mechanism for token errors in SD instead of relying on re-inference, thereby addressing the challenge of token errors and enabling acceleration.
(3) Third, we propose a kinematic-based strategy to dynamically rectify the acceptance threshold during inference, thereby solving the challenge of determining this threshold while maintaining a high success rate across multiple tasks.
\begin{figure*}[!t]
    \centering
    \includegraphics[width=7in]{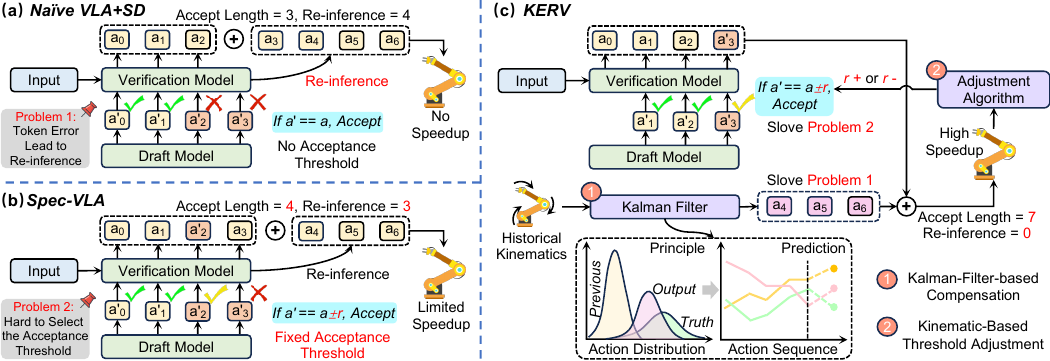}
    \vspace{-6mm}
	\caption{(a)~Naive VLA+SD \textit{vs.} (b)~SpecVLA \textit{vs.} (c)~The Proposed \textit{KERV}}
    \Description{fig1}
	\label{fig:1}
    \vspace{-5mm}
\end{figure*}
Moreover, we design a dedicated hardware mapping for the \textit{KERV} framework based on its computational characteristics and implement it on existing hardware platforms.
We evaluate \textit{KERV} across various specific environments (e.g., LIBERO~\cite{libero}), analyze its performance drivers, and discuss its optimal configurations. In summary, our contributions are below:
\begin{itemize}[leftmargin=*]
    \item[$\bullet$] We reveal the discrepancy between the token-domain (VLA models) and the kinematic-domain (conventional kinematic-based methods) in robot control, demonstrating the feasibility and advantages of their combination.
    \item[$\bullet$] Based on these, we take SD as the starting point and combine VLA models and kinematic-based prediction, proposing a kinematic-rectified SD framework termed \textit{KERV}. \textit{KERV} compensates for SD errors via kinematic-based action prediction and leverages kinematics as the metric to achieve adaptive acceptance threshold rectification.
    \item[$\bullet$] We implemented \textit{KERV} on existing hardware platforms (tailored to its computational characteristics) and validated the merits of the VLA-kinematics combination via extensive experiments. We believe \textit{KERV} will play a role in future embodied intelligence development and AI-physical combination.
\end{itemize}

We evaluate \textit{KERV} across diverse experiments and tasks. Experimental results show that compared with naive integration, \textit{KERV} achieves 1.48$\times$$\sim$1.57$\times$ acceleration, while maintaining the high success rate of VLA models. 
Additionally, versus SOTA works such as SpecVLA~\cite{spec-vla}, \textit{KERV} achieves 27\%$\sim$37\% speedup with almost no Success Rate loss. 

%% file: _tex/2_background.tex
\vspace{-2mm}
\section{Preliminaries}
\label{tex_background}

\vspace{-1mm}
\subsection{VLA Models}
\label{tex_background_1}

\noindent \textbf{Model Architecture.} 
Vision-Language-Action (VLA) models comprise three core components~\cite{openvla, rt2}: ViT encoders (converting visual data into tokenized representations), an LLM backbone (fusing multi-modal information and enabling reasoning), and an action de-tokenizer (decoding outputs into concrete actions).

\noindent \textbf{Action Decoding Paradigms.}
Each inference of VLA models produces one action slice $A_{i}$, requiring a sequence of action slices to complete a task~\cite{AMS, boosting-vla}. 
Each slice contains 7 DoF: position $X,Y,Z$ of the end gripper, joint rotation angles
$\theta_{X}, \theta_{Y}, \theta_{Z}$, and binary gripper control signal $G$, necessitating 7 autoregressive inference steps per slice.
Each DoF is encoded as a token $a_{j}$.
VLA models predict the most probable token $a_{j}$ based on the previously tokens $a_{0:j-1}$, visual observations $\mathbb{O}$, language prompts $\mathbb{P}$, and the learnable model parameters $\mathbb{W}$, as Eq.~\eqref{eq:2-1} shown:
\begin{equation}
a_{j} = \mathop{\arg\max}\limits_{a_{j}}  \big [ P(a_{j} \ | \ a_{0:j-1}, \mathbb{O}, \mathbb{P}, \mathbb{W}) \big ], (0\leq j \leq6).
\label{eq:2-1}
\end{equation}

\noindent \textbf{Acceleration for VLA Models.}
VLA acceleration works fall into three categories: model architecture innovation (such as RoboMamba~\cite{robomamba}, FastV~\cite{fastv}, TinyVLA~\cite{tinyvla} and Edge-VLA~\cite{edgevla}), model compression (such as MBQ~\cite{mbq} and QAIL~\cite{QAIL}), and runtime optimization (such as VLA-Cache~\cite{vla-cache}, SpecPrune-VLA~\cite{specprune-vla}, Efficient-VLA~\cite{efficient-vla}, Fastdrive-VLA~\cite{fastdrive-vla}, DeeR-VLA~\cite{deer-vla}, CEED-VLA~\cite{ceed-vla} and MoLe-VLA~\cite{mole-vla}).
However, despite favorable acceleration, existing work still leaves limitations in optimizing the VLA decoding paradigm, requiring decoding optimizations such as SD.

\vspace{-1.5mm}
\subsection{Speculative Decoding}
\label{tex_background_2}

\noindent \textbf{SD for LLMs.}
SD comprises a draft model $M_{\textnormal{D}}$ and a verification model $M_{\textnormal{V}}$.
In the draft phase, $M_{\textnormal{D}}$ predicts multiple tokens using hidden states $f_{1:t}$ and embeddings $e_{0:t}$ , as shown in Eq.~\eqref{eq:2-2}.
For simplicity, we use $a_{t+1:j-1}$ to denote both embeddings and hidden features.
In the verification phase, $M_{\textnormal{V}}$ ensures draft token quality; in greedy search, draft token $\hat{a_{j}}$ is accepted only if strictly matching $M_{\textnormal{V}}$'s predicted token $a_{j}$, as shown in Eq.~\eqref{eq:2-3}:
\begin{equation}
\textnormal{Draft:} \ \hat{a}_{j} = M_{\textnormal{D}}(f_{1:t}, e_{0:t}, \hat{a}_{t+1:j-1}).
\label{eq:2-2}
\end{equation}
\begin{equation}
\textnormal{Verify:} \ a_{j} = M_{\textnormal{V}}(a_{0}\sim \hat{a}_{j-1}, \mathbb{P}, \mathbb{W}), \
\begin{aligned}
\begin{cases}
\ \textnormal{If} \ a_{j} =\hat{a}_{j}, \textnormal{Accept}. \\
\ \textnormal{If} \ a_{j} \neq \hat{a}_{j}, \textnormal{Resample}.
\end{cases} 
\end{aligned}
\label{eq:2-3}
\end{equation}

SD has evolved in three stages: pioneering works (e.g., Medusa~\cite{medusa} and Medusa-CTC~\cite{medusa-CTC}) introduced multi-head parallel generation with tree-attention verification; evolutionary frameworks (e.g., EAGLE series~\cite{eagle, eagle-2}) enhanced draft modeling for better quality and speedups; recent studies (e.g., EAGLE-3~\cite{eagle-3} and HASS~\cite{HASS}) integrated training-time testing to further boost capabilities.

\noindent \textbf{SD for VLA Models.}
However, integrating SD with VLA models is non-trivial. Naive integration (shown in Fig.~\ref{fig:1}(a)) faces two key challenges: (1) SD relies on computationally expensive re-inference to address token errors, which is high-cost. (2) To mitigate token errors, the acceptance threshold in SD requires careful adjustment.
While Spec-VLA~\cite{spec-vla} explores the determination of acceptance threshold (shown in Fig.~\ref{fig:1}(b)), it still relies on re-inference to mitigate errors, which limits the speedup. 
Furthermore, it adopts a fixed acceptance threshold, overlooking the complexity and dynamics of various embodied intelligence environments.

\vspace{-1.5mm}
\subsection{Kinematic-Based Robot Control}
\label{tex_background_3}
\noindent \textbf{Principles of Kinematic-Based Methods.}
Traditional robotic control~\cite{kinematics-1, kinematics-2} takes kinematic information (e.g., joint angles, link displacements, motion velocities) as input to directly regulate robot movements, with no token representation involved.
It centers on pre-defined mathematical models and control laws, which require human design. Moreover, it relies on accurate robot dynamic modeling and real-time sensor feedback to achieve precise trajectory tracking and motion execution.
Unlike VLA models, it lacks the capability for long-term task planning or environmental perception.

\noindent \textbf{Kalman-Filter-Based Robot Control.}
The Kalman Filter (KF) is a classic technique in traditional control methods~\cite{kalman-1, kalman-2}.
It updates control signals based on the gaps between the prior action distribution and the true value.
In robotic control, it leverages kinematic constraints to predict future action states, achieving high accuracy in most scenarios.
This lightweight, real-time optimization aligns with the efficiency requirements of kinematic-based control, solidifying its role as a staple for precise action regulation.

%% file: _tex/3_motivation.tex
\vspace{-1.5mm}
\section{Token-Kinematic Discrepancy}
\label{tex_motivation}

In this section, we analyze SD-induced errors and the associated acceptance threshold, identify the discrepancy between the token-domain and kinematic-domain paradigms, and demonstrate the insights of combining VLA models with kinematic-based methods.

\begin{table}[!t]
    \centering
    \caption{Case Studies on Naive Integration of VLA and SD \vspace{-3mm}}
    \label{tab:3-1}
    \scriptsize
    \begin{tabular}{c|cccc|ccccc}
    \toprule
    \toprule
    \multirow{2}{*}{\textbf{Env.}} & \multicolumn{4}{c|}{\textbf{AR}} & \multicolumn{5}{c}{\textbf{Naive VLA+SD}} \\
    \cmidrule{2-10}
    ~ & \textbf{SR} & \textbf{Speed} & \textbf{Step} & \textbf{T(s)} & \textbf{SR} & \textbf{Speed} & \textbf{Step} & \textbf{\textit{T}(s)} & \textbf{AFEP} \\
    \midrule
    Goal & 77.0\% & 1$\times$ & 157.6 & 0.188 & 76.2\% & 0.86$\times$ & 159.2 & 0.217 & 2.04 \\
    Object & 71.2\% & 1$\times$ & 191.7 & 0.197 & 68.6\% & 0.96$\times$ & 195.9 & 0.200 & 1.75 \\
    Spatial & 82.8\% & 1$\times$ & 126.9 & 0.198 & 82.8\% & 0.98$\times$ & 127.3 & 0.201 & 1.59 \\
    Long & 54.4\% & 1$\times$ & 393.2 & 0.190 & 50.2\% & 0.91$\times$ & 400.7 & 0.217 & 1.67 \\
    \bottomrule
    \bottomrule
    \end{tabular}
    \vspace{-5mm}
\end{table}

\begin{figure}[!b]
	\centering
    \vspace{-4mm}
    \includegraphics[width=3.3in]{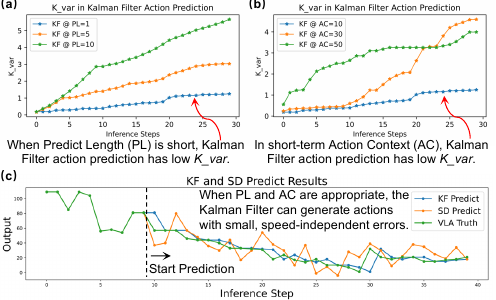}
    \vspace{-4mm}
	\caption{Discrepancy about Errors}
    \Description{fig2}
	\label{fig:2}
\end{figure}

\vspace{-2mm}
\subsection{Discrepancy about Errors}
\label{tex_motivation_1}

\textit{Motivation \ding{172}: SD relies on high-cost re-inference to address generation errors, hindering inference speed. In contrast, kinematic methods enable fast generation with small, speed-independent errors.}

\noindent \textbf{Token-Domain Analysis about Generation Errors.}
We naively adapted SD via the EAGLE framework~\cite{eagle} and OpenVLA~\cite{openvla} model, evaluating Success Rate (SR), speed, average inference steps, and per-step latency (\textit{T}). 
We further report SD’s Average First Error Position (AFEP).
The results in Tab.~\ref{tab:3-1} show that prediction errors (AFEP 1.59$\sim$2.04) force draft token abandonment and subsequent re-inference, leading to lower speed than AR (0.188s$\sim$0.198s v.s. 0.200s$\sim$0.217s).
Clearly, acceleration hinges on properly addressing SD-generated errors with a lower-cost compensation approach instead of re-inference.

\noindent \textbf{Kinematic-Domain Analysis about Prediction Errors.} Fortunately, Kalman Filter (KF) features ultra-low computational overhead and fast action prediction, making them ideal alternatives to re-inference.
While KF also incurs certain errors, so we propose $K_{\textnormal{var}}$ to characterize the kinematic variability, as shown in Eq.~\eqref{eq:3-1}:
\begin{equation}
K_{\textnormal{var}} = \sum_{step} \big \| action_{j}^\textnormal{correct} - action_{j}^{\textnormal{error}} \big \|_{1}, \ (0 \leq j \leq 6),
\label{eq:3-1}
\end{equation}
The Action Context (AC) and Predict Length (PL) will affect KF's prediction errors.
We first test KF under various Prediction Length (PL) and the results are shown in Fig.~\ref{fig:2}~(a).
When KF only predicting 1 step ahead (@ PL=1), it achieve minimal $K_{\textnormal{var}}$.
When PL increases, the actions predicted by KF become imprecise.
Additionally, we test how action context (AC) impacts KF performance.
As Fig.~\ref{fig:2}~(b) shows, KF's $K_{\textnormal{var}}$ is minimized in short-term AC (@ AC=10).
Beyond that, the $K_{\textnormal{var}}$ increases as the AC grows. 
As Fig.~\ref{fig:2}(c) shows, with appropriate PL and AC, KF outperforms SD’s draft model in action accuracy, and its errors do not impact speed.

\noindent \textit{Insight \ding{172}: KF can be employed to predict actions for compensating SD errors, thereby avoiding computationally expensive re-inference and enabling end-to-end acceleration.}

\begin{figure}[!b]
	\centering
    \vspace{-4mm}
    \includegraphics[width=3.3in]{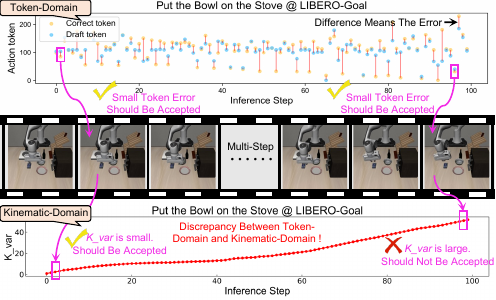}
    \vspace{-4mm}
	\caption{Discrepancy about Acceptance Threshold}
    \Description{fig3}
	\label{fig:3}
\end{figure}

\vspace{-2mm}
\subsection{Discrepancy about Acceptance Threshold}
\label{tex_motivation_2}

\textit{Motivation \ding{173}: Existing work adopts a fixed relaxed acceptance threshold for SD errors. In contrast, from a kinematic perspective, some accepted errors should be discarded, and the acceptance threshold should be dynamically adjusted.}

\noindent \textbf{Token-Domain Analysis about Acceptance Threshold.}
Existing work like Spec-VLA~\cite{spec-vla} adopts a fixed relaxed acceptance threshold, forcing the VLA model to accept draft tokens with errors less than $r$ ($r$ means the relaxed threshold).
However, determining $r$ for diverse embodied tasks and environments remains challenging.
Experiments show that a large $r$ will increase inference steps and even cause task failure, and a small $r$ will lead to limited speedup.
Furthermore, even for a specific task, a fixed threshold is irrational: as errors accumulate, the more inference steps there are, the more erroneous tokens that should no longer be accepted.

\noindent \textbf{Kinematic-Domain Analysis about Acceptance Threshold.}
From a kinematic perspective, we analyze the relaxed acceptance threshold.
Specifically, we record the SD errors and corresponding $K_{\textnormal{var}}$ during inference (as shown in Fig.~\ref{fig:3}).
We select two representative time steps. 
At the first case step, the SD-generated token error is small (error=2) and would be accepted under Spec-VLA’s relaxed acceptance threshold ($r$=9).
At this point, $K_{\textnormal{var}}$ is small, indicating this error is acceptable.
At the second case step, the SD-generated token error remains small (error=1) and would also be accepted under this fixed threshold. 
However, the corresponding $K_{\textnormal{var}}$ is large, requiring rejection of this erroneous token, revealing a token-kinematic discrepancy in the acceptance threshold.
This indicates that fixed thresholds perform poorly, showing the inadequacy of determining such thresholds solely in the token domain.

\noindent \textit{Insight \ding{173}: The acceptance threshold determined solely from the token domain is suboptimal and demands dynamic adjustment, while kinematic variability serves as a suitable metric to guide this process.}

%% file: _tex/4_method.tex
\vspace{-2.5mm}
\section{\textit{KERV} Framework}
\label{tex_method}

The aforementioned analysis and insights drive us to combine VLA models and conventional kinematic-based methods. 
In this section, we will detail how we combine them and leverage them to address the two key challenges of SD within the \textit{KERV} framework.

\begin{figure}[!t]
	\centering
    \includegraphics[width=3.3in]{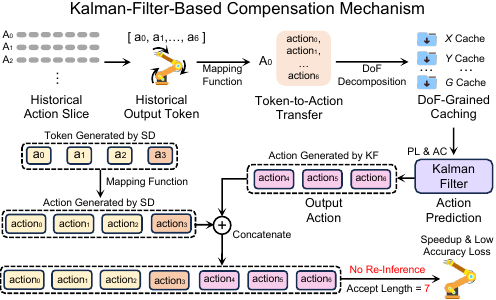}
    \vspace{-4mm}
	\caption{KF-Based Compensation Mechanism in \textit{KERV}}
    \vspace{-6mm}
    \Description{fig4}
	\label{fig:4}
\end{figure}

\vspace{-2mm}
\subsection{\textbf{KF-Based Compensation Mechanism}}
\label{tex_method_1}

From Insight \ding{172}, to address the challenge of re-inference when facing token errors in SD, we combine kinematic-based methods and VLA models, proposing a compensation mechanism.
As Fig.~\ref{fig:4} illustrates, the core idea is that when an SD error occurs, we use KF to complete action prediction for the remainder of the current action slice instead of re-inference, thereby achieving acceleration.

We first build a mapping function $Map(\cdot, \cdot)$ of the VLA model's token-action transfer, as shown in Eq.~\eqref{eq:4-1}.
$Map(\cdot)$ takes the output token $A_{i}$ and the pre-sampled action distribution $D_{action}^{\textnormal{Norm-key}}$ of the robotic arm as input, and outputs the corresponding action signals.
Norm-key is an inherent statistic of the robotic arm.
\begin{equation}
action_{j} = Map\big ( (A_{i} \ | \ \mathbb{O}, \mathbb{P}, \mathbb{W}) ; \ D_{action}^{\textnormal{Norm-key}} \big ), \ (0 \leq j \leq 6).
\label{eq:4-1}
\end{equation}

Then, we build caching for these actions in DoF granularity, represented as $\textnormal{Cache}_{X}, \textnormal{Cache}_{Y}, \cdots, \textnormal{Cache}_{G}$.
We use these caches as the historical input sequence required by KF to enable accurate action prediction, as shown in Eq.~\eqref{eq:4-2}. 
$\mathcal{I},\mathcal{M}, \mathcal{P}$ is the initial parameters of KF.
$p$ means the first error position of SD.
We set the predicted length (PL) to 1 and the action context (AC) to 10.
\begin{equation}
Kalman^{PL} \big ( \textnormal{Cache}_{X\sim G}^{AC}; (\mathcal{I}, \mathcal{M}, \mathcal{P})\big ) = 
\begin{cases}
action_{0\sim p}^{\textnormal{KF}},  \textnormal{Discard} \\
action_{p\sim 6}^{\textnormal{KF}},  \textnormal{Keep}
\end{cases} 
\label{eq:4-2}
\end{equation}

After that, we concatenate the actions from SD ($action_{0\sim p}^{\textnormal{SD}}$) and the actions generated by KF ($action_{p\sim6}^{\textnormal{KF}}$) to form a complete action slice, and send it to the robotic arm.
In the next step, the VLA model re-encodes environmental information and re-prefills for the next generation. 
Thus, the proposed compensation mechanism does not require modifying the KV Cache.
To maintain a short PL for KF’s prediction accuracy, we activate the compensation mechanism intermittently.
Therefore, after each time of compensation, we use SD for the next $n$ steps with KF disabled.
We set $n=4$ and discuss it in Section~\ref{tex_experiments}.

\begin{figure}[!t]
	\centering
    \includegraphics[width=3.3in]{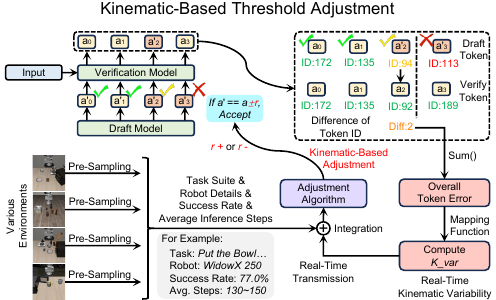}
    \vspace{-4mm}
	\caption{Kinematic-Based Threshold Adjustment in \textit{KERV}}
    \vspace{-6mm}
    \Description{fig5}
	\label{fig:5}
\end{figure}

\vspace{-2mm}
\subsection{\textbf{Kinematic-Based Threshold Adjustment}}
\label{tex_method_2}
\noindent 
From Insight \ding{173}, to address the challenge of determining the acceptance threshold, we propose a kinematic-based threshold adjustment strategy.
As shown in Fig.~\ref{fig:5}, we compute the ID difference between true tokens and those accepted via relaxed thresholds.
Note that unaccepted error tokens do not participate in the calculation.
Then, we aggregate the differences at each step and convert them to kinematic variability through the mapping function to calculate $K_{\textnormal{var}}$.
Moreover, we pre-sample in diverse environments to gather the necessary information, including the task suite, robot details, success rate, and average inference steps.

After that, we use this information to determine the appropriate maximal and minimal thresholds $r_{\textnormal{max}}$ and $r_{\textnormal{min}}$, as well as the appropriate algorithm parameters $\tau$ and $\varphi$, and build them into look-up tables (for most tasks, we set $r_{\textnormal{max}}$=15 and $r_{\textnormal{min}}$=5).
We integrate these lookup tables and kinematic variability, then feed them into an adjustment algorithm that takes them as inputs and outputs acceptance threshold adjustment guidance.
The details of the adjustment algorithm are shown in Alg.~\ref{alg:1}.
This kinematic-based strategy can automatically adjust the acceptance threshold at runtime, enabling better preservation of the VLA model’s high success rate during the SD process.
\vspace{-2.5mm}
\begin{algorithm}[!b]
\caption{Adjustment Algorithm}
\footnotesize
\begin{algorithmic}[1]
\Require Kinematic Variability $K_{\textnormal{var}}$, Acceptance Threshold Maximum and Minimum $r_{\textnormal{max}}, r_{\textnormal{min}}$, Parameters Look-up Table Generated by Pre-Sampling $[\tau_{1}, \tau_{2}, \cdots,\tau_{n}]$, $[\varphi_{1}, \varphi_{2},\cdots, \varphi_{n}]$.
\Ensure Task $T$, Robot $Robot$, Success Rate $R$ and Average Inference Steps $S$
\State Searching $\tau$ and $\varphi$ in $[\tau_{1}, \cdots,\tau_{n}]$ and $[\varphi_{1}, \cdots, \varphi_{n}]$ based on [$T$, $Robot$, $R$, $S$]
\If {$\tau \in [\tau_{1}, \tau_{2}, \cdots,\tau_{n}]$ and $\varphi \in [\varphi_{1}, \varphi_{2},\cdots, \varphi_{n}]$}
    \For{Each Step $t$} Compute $\Delta K_{\textnormal{var}}^{t} = K_{\textnormal{var}}^{t} - K_{\textnormal{var}}^{t-1}$ 
    \If{$\Delta K_{\textnormal{var}} \neq 0$} 
    \State $\Delta r^{t} = (r_{\textnormal{max}}-r_{\textnormal{min}})*exp((-\frac{\Delta K_{\textnormal{var}}^{t}}{K_{\textnormal{var}}^{S}})^{\varphi})$ 
    \Else $\ \textnormal{\textbf{continue}}$
    \EndIf
    \State Update Acceptance Threshold: $ r^{t+1} \leftarrow r^{t} +\Delta r^{t}$
    \If{$r^{t+1} \leq r_{\textnormal{min}}$} \textbf{break}
    \EndIf
    \EndFor
\EndIf
\end{algorithmic}
\label{alg:1}
\end{algorithm}

%% file: _tex/5_implementation.tex
\vspace{-1mm}
\section{System Implementation}
\label{tex_implementation}

This section explores the efficient implementation of \textit{KERV} on existing hardware platforms from a computational perspective based on CPU-GPU collaboration.

\begin{figure}[!t]
	\centering
    \includegraphics[width=3.3in]{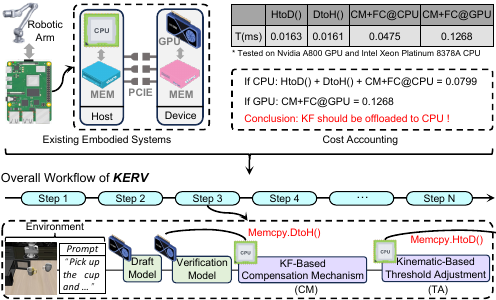}
    \vspace{-4mm}
	\caption{System Implementation of \textit{KERV}}
    \Description{fig6}
	\label{fig:6}
    \vspace{-5mm}
\end{figure}

\vspace{-2mm}
\subsection{GPU-Side Draft and Verify}
\label{tex_implementation_1}
Both the draft model and verification model involve substantial computation.
Measurements show the draft model requires 0.07 GFLOPs per inference, while the verification model consumes 3.92 TFLOPs per inference. 
We therefore adopt GPU acceleration for their inference processes.
Owing to the draft model’s smaller size, its GPU memory footprint is significantly lower than that of the verification model.
The draft model incurs a memory overhead of 700MB, while the 7B verification model consumes 15GB of memory, leading to a negligible footprint for the former.

\vspace{-2mm}
\subsection{\textbf{CPU-Side Compensation and Adjustment}}
\label{tex_implementation_2}
In \textit{KERV}, the KF-Based compensation mechanism and the kinematic-based threshold adjustment involve extensive logical decisions but small FLOPs, which suggests they may be more efficient on the CPU than on the GPU.
However, CPU execution requires multiple CPU-GPU data transfers, introducing extra overhead, which prompted detailed benchmarking.
As shown in Fig.~\ref{fig:6}, the existing embodied system integrates both CPU and GPU. 
On an Nvidia A800 GPU and an Intel Xeon Platinum 8378A CPU, we measured Host-to-Device (HtoD) and Device-to-Host (DtoH) memory copy latencies, implemented two parts, and quantified their latencies. 
Even with additional memory copy overhead, CPU-executed latency is still lower than that of the GPU.
Thus, for speed optimization, we offload these two components to the CPU.

\vspace{-3mm}
\subsection{Overall Workflow of \textit{KERV}}
\label{tex_implementation_3}
\textit{KERV} is implemented in $\sim$5000 lines of code. 
Fig.~\ref{fig:6} outlines its workflow: the environmental visual and textual prompt is encoded as input to the draft model, which autoregressively outputs tokens. 
The verification model evaluates these tokens with a relaxed acceptance threshold.
The verification model evaluates these tokens using a relaxed acceptance threshold. The results are then transferred to the CPU via DtoH memory copy for KF-based compensation and kinematic-based threshold adjustment, before being sent back to the GPU via HtoD memory copy for the next step.

%% file: _tex/6_experiment.tex
\vspace{-2mm}
\section{Experiments}
\label{tex_experiments}

\vspace{-1mm}
\subsection{\textbf{Setup}}
\label{tex_experiments_1}
Following OpenVLA~\cite{openvla}, we tested \textit{KERV} framework on the LIBERO benchmark~\cite{libero}.
We utilize four task suites, including LIBERO-Object, LIBERO-Spatial, LIBERO-Goal, and LIBERO-Long.
Each suite contains 10 tasks. 
For each task, we conduct 50 trials for testing.
We utilize the finetuned OpenVLA as the verification models and build a single LLaMA block~\cite{llama} as a draft model.
We trained the draft models based on the DeepSpeed~\cite{deepspeed} framework.
The training process was completed in 12 hours using 2$\times$ NVIDIA A800 (40G) GPUs.
We inherit the tree decoding mechanism from Eagle-2~\cite{eagle-2}. 
For tree-decoding, we set the maximum nodes to 50, the tree depth to 4, and used the top 8 tokens to construct the draft tree.
Given that Spec-VLA~\cite{spec-vla} is the only existing work integrating VLA and SD, we adopt it alongside naive VLA+SD as baselines.
We select an Nvidia A800 GPU and an Intel Xeon Platinum 8378A CPU as the testing hardware platform.

\begin{figure*}[!t]
    \centering
    \includegraphics[width=6.7in]{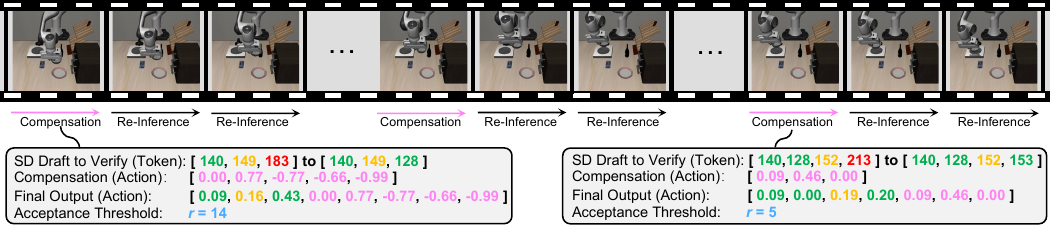}
    \vspace{-3.5mm}
	\caption{An Experimental Example of \textit{KERV} on LIBERO-Goal Environment}
    \Description{fig7}
	\label{fig:7}
    \vspace{-4mm}
\end{figure*}

\vspace{-2mm}
\subsection{\textbf{Evaluation Results}}
\label{tex_experiments_2}
\noindent \textbf{End-to-End Results.}
We report the Success Rate (SR), Speedup, Average First Error Position (AFEP), and average inference steps, as Tab.~\ref{tab:6-1}.
The results show that compared with naive combination, \textit{KERV} achieves 1.48$\times$$\sim$1.57$\times$ speedup, without SR loss.
Compared with Spec-VLA under various acceptance thresholds ($r$=9/15/20), \textit{KERV} achieves 27\%$\sim$37\% acceleration, with only 1.5\% and 0.4\% SR decrease on two environments.
Moreover, in all four environments, \textit{KERV} has the smallest average inference steps.
All these results demonstrate that combining kinematics with VLA models enables a better speed-accuracy Pareto Frontier.

\begin{table}[!b]
    \centering
    \vspace{-5mm}
    \caption{End-to-End Results of \textit{KERV} and Baselines \vspace{-3mm}}
    \label{tab:6-1}
    \scriptsize
    \begin{tabular}{ccccccc}
    \toprule
    \toprule
    {\textbf{Env.}} & \textbf{Method} & {\textbf{SR}} & \textbf{Speed} & \textbf{AFEP} & \textbf{Steps} & \textbf{Hardware}\\
    \midrule
    \multirow{5}{*}{Goal} & \cellcolor{gray!20}{Naive VLA+SD} & \cellcolor{gray!20}{76.2\%} & \cellcolor{gray!20}{1.00$\times$} & \cellcolor{gray!20}{2.04} & \cellcolor{gray!20}{159.2} & \cellcolor{gray!20}{GPU} \\
    ~ & \cellcolor{yellow!20}{SpecVLA ($r$=9)} & \cellcolor{yellow!20}{75.4\%} & \cellcolor{yellow!20}{1.19$\times$} & \cellcolor{yellow!20}{2.97} & \cellcolor{yellow!20}{157.3} & \cellcolor{yellow!20}{GPU} \\
    ~ & \cellcolor{yellow!20}{SpecVLA ($r$=15)} & \cellcolor{yellow!20}{71.0\%} & \cellcolor{yellow!20}{1.23$\times$} & \cellcolor{yellow!20}{3.63} & \cellcolor{yellow!20}{166.8} & \cellcolor{yellow!20}{GPU} \\
    ~ & \cellcolor{yellow!20}{SpecVLA ($r$=20)} & \cellcolor{yellow!20}{64.2\%} & \cellcolor{yellow!20}{1.21$\times$} & \cellcolor{yellow!20}{4.23} & \cellcolor{yellow!20}{176.3} & \cellcolor{yellow!20}{GPU} \\
    ~ & \cellcolor{green!20}{\textit{KERV}} & \cellcolor{green!20}{\textbf{75.6\%}} & \cellcolor{green!20}{\textbf{1.54$\times$}} & \cellcolor{green!20}{\textbf{4.73}} & \cellcolor{green!20}{\textbf{153.5}} & \cellcolor{green!20}{CPU+GPU} \\
    \cmidrule{1-7}
    \multirow{5}{*}{Object} & \cellcolor{gray!20}{Naive VLA+SD} & \cellcolor{gray!20}{68.6\%} & \cellcolor{gray!20}{1.00$\times$} & \cellcolor{gray!20}{1.75} & \cellcolor{gray!20}{195.9} & \cellcolor{gray!20}{GPU} \\
    ~ & \cellcolor{yellow!20}{SpecVLA ($r$=9)} & \cellcolor{yellow!20}{70.0\%} & \cellcolor{yellow!20}{1.09$\times$}  & \cellcolor{yellow!20}{3.25} & \cellcolor{yellow!20}{200.0} & \cellcolor{yellow!20}{GPU}\\
    ~ & \cellcolor{yellow!20}{SpecVLA ($r$=15)} & \cellcolor{yellow!20}{62.4\%} & \cellcolor{yellow!20}{1.10$\times$} & \cellcolor{yellow!20}{3.91} & \cellcolor{yellow!20}{214.5} & \cellcolor{yellow!20}{GPU} \\
    ~ & \cellcolor{yellow!20}{SpecVLA ($r$=20)} & \cellcolor{yellow!20}{58.0\%} & \cellcolor{yellow!20}{1.10$\times$} & \cellcolor{yellow!20}{4.18} & \cellcolor{yellow!20}{221.5} & \cellcolor{yellow!20}{GPU} \\
    ~ & \cellcolor{green!20}{\textit{KERV}} & \cellcolor{green!20}{\textbf{72.3\%}} & \cellcolor{green!20}{\textbf{1.49$\times$}} & \cellcolor{green!20}{\textbf{4.71}} & \cellcolor{green!20}{\textbf{186.8}} & \cellcolor{green!20}{CPU+GPU} \\
    \cmidrule{1-7}
    \multirow{5}{*}{Spatial} & \cellcolor{gray!20}{Naive VLA+SD} & \cellcolor{gray!20}{82.8\%} & \cellcolor{gray!20}{1.00$\times$} & \cellcolor{gray!20}{1.59} & \cellcolor{gray!20}{127.3} & \cellcolor{gray!20}{GPU} \\
    ~ & \cellcolor{yellow!20}{SpecVLA ($r$=9)} & \cellcolor{yellow!20}{\textbf{85.2\%}} & \cellcolor{yellow!20}{1.25$\times$}  & \cellcolor{yellow!20}{3.27} & \cellcolor{yellow!20}{124.9} & \cellcolor{yellow!20}{GPU} \\
    ~ & \cellcolor{yellow!20}{SpecVLA ($r$=15)} & \cellcolor{yellow!20}{80.4\%} & \cellcolor{yellow!20}{1.26$\times$} & \cellcolor{yellow!20}{3.80} & \cellcolor{yellow!20}{128.7} & \cellcolor{yellow!20}{GPU} \\
    ~ & \cellcolor{yellow!20}{SpecVLA ($r$=20)} & \cellcolor{yellow!20}{77.8\%} & \cellcolor{yellow!20}{1.24$\times$} & \cellcolor{yellow!20}{4.14} & \cellcolor{yellow!20}{133.3} & \cellcolor{yellow!20}{GPU} \\
    ~ & \cellcolor{green!20}{\textit{KERV}} & \cellcolor{green!20}{83.7\%} & \cellcolor{green!20}{\textbf{1.57$\times$}} & \cellcolor{green!20}{\textbf{4.67}} & \cellcolor{green!20}{\textbf{120.9}} & \cellcolor{green!20}{CPU+GPU} \\
    \cmidrule{1-7}
    \multirow{5}{*}{Long} & \cellcolor{gray!20}{Naive VLA+SD} & \cellcolor{gray!20}{50.2\%} & \cellcolor{gray!20}{1.00$\times$} & \cellcolor{gray!20}{1.67} & \cellcolor{gray!20}{400.7} & \cellcolor{gray!20}{GPU} \\
    ~ & \cellcolor{yellow!20}{SpecVLA ($r$=9)} & \cellcolor{yellow!20}{\textbf{49.2\%}} & \cellcolor{yellow!20}{1.12$\times$} & \cellcolor{yellow!20}{2.82} & \cellcolor{yellow!20}{408.9} & \cellcolor{yellow!20}{GPU} \\
    ~ & \cellcolor{yellow!20}{SpecVLA ($r$=15)} & \cellcolor{yellow!20}{36.2\%} & \cellcolor{yellow!20}{1.13$\times$} & \cellcolor{yellow!20}{3.63} & \cellcolor{yellow!20}{439.6} & \cellcolor{yellow!20}{GPU} \\
    ~ & \cellcolor{yellow!20}{SpecVLA ($r$=20)} & \cellcolor{yellow!20}{25.8\%} & \cellcolor{yellow!20}{1.10$\times$} & \cellcolor{yellow!20}{4.13} & \cellcolor{yellow!20}{464.6} & \cellcolor{yellow!20}{GPU}\\
    ~ & \cellcolor{green!20}{\textit{KERV}} & \cellcolor{green!20}{48.8\%} & \cellcolor{green!20}{\textbf{1.48$\times$}} & \cellcolor{green!20}{\textbf{4.64}} & \cellcolor{green!20}{\textbf{391.2}} & \cellcolor{green!20}{CPU+GPU} \\
    \bottomrule
    \bottomrule
    \end{tabular}
\end{table}

\noindent \textbf{Example Presentation.}
We present an example to illustrate KERV’s details in the SD process.
As shown in Fig.~\ref{fig:7}, the draft model generates tokens [140, 149, 183] for the first case step: yellow number 149 denotes an erroneous token accepted by the relaxed threshold, green number 140 a fully correct token, and red number 183 an erroneous token rejected for exceeding the threshold.
After verification, the verification model corrects 183 to 128, and these tokens are decoded into actions (action scope -1$\sim$1). 
The Compensation Mechanism (hereinafter referred to as CM) compensates for residual positions using predicted actions (pink numbers), with the final output being the concatenation of SD-generated actions and CM-predicted actions.
At this step, the acceptance threshold $r$ is set to 14 under the Threshold Adjustment (TA) strategy. 
For the second case step, $r$ is adjusted to 5, demonstrating TA’s dynamic effect.

\noindent \textbf{Ablation Studies.}
We performed ablation experiments on two components of \textit{KERV}: CM and TA, and the results are shown in Tab.~\ref{tab:6-3}. 
The four environments show the same trend. 
With only the CM, the system achieves high speed. 
However, as the threshold remains unadjusted, it accepts more erroneous SD-generated tokens, leading to significant accuracy loss. 
In contrast, with only TA, the system automatically adjusts the threshold based on dynamic changes, thus yielding favorable accuracy. 
Without CM, however, re-inference is unavoidable, resulting in limited acceleration.

\begin{table}[!b]
    \centering
    \vspace{-4mm}
    \caption{Ablation Studies of \textit{KERV} \vspace{-3.5mm}}
    \label{tab:6-2}
    \scriptsize
    \begin{tabular}{cc|cc|cc|cc|cc}
    \toprule
    \toprule
    \multicolumn{2}{c|}{\textbf{\textit{KERV}}} & \multicolumn{2}{c|}{\textbf{Goal}} & \multicolumn{2}{c|}{\textbf{Object}} & \multicolumn{2}{c|}{\textbf{Spatial}} & \multicolumn{2}{c}{\textbf{Long}} \\
    \cmidrule{3-10}
    \textbf{CM} & \textbf{TA} & \textbf{SR} & \textbf{Speed} & \textbf{SR} & \textbf{Speed} & \textbf{SR} & \textbf{Speed} & \textbf{SR} & \textbf{Speed} \\
    \midrule
    \textcolor{green}{\CheckmarkBold} & \textcolor{red}{\XSolidBrush} & 72.6\% & 1.57$\times$ & 68.1\% & 1.48$\times$ & 80.9\% & 1.59$\times$ & 46.9\% & 1.48$\times$ \\
    \textcolor{red}{\XSolidBrush} & \textcolor{green}{\CheckmarkBold} & 77.0\% & 1.17$\times$ & 72.9\% & 1.07$\times$ & 85.0\% & 1.21$\times$ & 49.6\% & 1.11$\times$ \\
    \cmidrule{3-10}
    \textcolor{green}{\CheckmarkBold} & \textcolor{green}{\CheckmarkBold} & 75.6\% & 1.54$\times$ & 72.3\% & 1.49$\times$ & 83.7\% & 1.57$\times$ & 48.8\% & 1.48$\times$ \\
    \bottomrule
    \bottomrule
    \end{tabular}
\end{table}

\vspace{-2mm}
\subsection{Discussion}
\label{tex_experiments_4}
\begin{figure}[!t]
    \centering
    \includegraphics[width=3in]{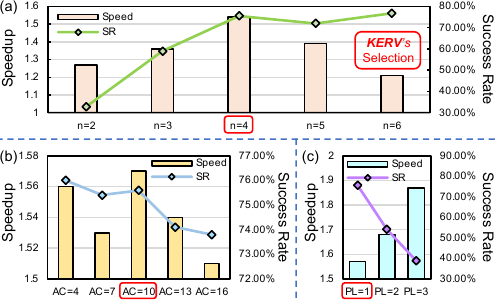}
    \vspace{-4mm}
	\caption{Experiments of Hyper-Parameters in \textit{KERV}}
    \Description{fig8}
	\label{fig:8}
    \vspace{-5mm}
\end{figure}

\noindent \textbf{Hyperparameters.}
The hyperparameters of \textit{KERV} include the prediction length (PL), the action context (AC) of KF, and the number of SD steps ($n$) executed after each time of KF. 
We evaluate \textit{KERV}'s performance across different hyperparameters on LIBERO-Goal, with the results presented in Fig.~\ref{fig:8}.
From Fig.~\ref{fig:8}~(a), when $n$<4, the SR decreases rapidly.
When $n$>4, the acceleration ratio decreases gradually due to the increase in SD steps, and the difference in accuracy is not large.
Therefore, $n$=4 is the optimal choice.
From Fig.~\ref{fig:8}~(b), as AC varies, the speed fluctuates slightly, while SR decreases with increasing AC.
Thus, we use AC=10 as the basis selection for \textit{KERV}.
From Fig.~\ref{fig:8}~(c), as the PL increases, SR dropped sharply. 
This is because KF can only maintain accurate predictions in a short context, and when the context expands, it is equivalent to making the next prediction with the predicted value with error, resulting in a sharp decrease in SR.
Therefore, we select PL=1 in \textit{KERV}.

\noindent \textbf{Hardware Implementation Analysis.}
We test the end-to-end time overhead on different hardware platforms in four environments. 
The Results are shown in Tab.~\ref{tab:6-3}.
Compared with GPU-only, our proposed CPU+GPU implementation achieves lower end-to-end latency, showing 7\%$\sim$13\% acceleration contribution.

\noindent \textbf{Generability.}
Due to the limited generalization of the VLA model itself, it is difficult to adapt to a variety of real-world robotic arms.
Therefore, we do not do real machine adaptation, but pay more attention to how to combine the VLA model with kinetic methods. 

\begin{table}[!b]
    \centering
    \vspace{-5mm}
    \caption{Hardware Implementation Analysis of \textit{KERV} \vspace{-3mm}}
    \label{tab:6-3}
    \scriptsize
    \begin{tabular}{c|c|c|c|c}
    \toprule
    \toprule
    \textbf{Detials} & \textbf{Goal} & \textbf{Object} & \textbf{Spatial} & \textbf{Long} \\
    \midrule
    \textit{KERV} on GPU @ 50 Trials & 12473.4 s & 14073.6 s & 9111.8 s & 31308.7 s \\
    \textit{KERV} on CPU+GPU @ 50 Trials & 11206.6 s & 13147.9 s & 8183.2 s & 27607.1 s \\
    \cmidrule{2-5}
    Acceleration of CPU+GPU & 1.11$\times$ & 1.07$\times$ & 1.11$\times$ & 1.13$\times$ \\
    \bottomrule
    \bottomrule
    \end{tabular}
\end{table}

%% file: _tex/7_conclusion.tex
\vspace{-1.5mm}
\section{Conclusion}
\label{tex_conclusion}

In this paper, we first analyze the discrepancy between token-domain and kinematic-domain to derive insights. 
Then we target SD optimization and propose \textit{KERV} framework, which contains a Kalman-Filter-based compensation mechanism and a kinematic-based threshold adjustment strategy.
Experimental results show that \textit{KERV} achieves 27\%$\sim$37\% acceleration with nearly no SR loss.

%% file: _tex/8_ack.tex
\section{Acknowledgment}
This paper is supported by National Natural Science Foundation of China (NSFC) Projects (No. 62227809).